# Possibilistic logic bases and possibilistic graphs

**Salem BENFERHAT - Didier DUBOIS - Laurent GARCIA - Henri PRADE**

Institut de Recherche en Informatique de Toulouse (I.R.I.T.)
Université Paul Sabatier — 118, route de Narbonne
31062 Toulouse — France
{benferhat, dubois, garcia, prade}@irit.fr

## Abstract

Possibilistic logic bases and possibilistic graphs are two different frameworks of interest for representing knowledge. The former stratifies the pieces of knowledge (expressed by logical formulas) according to their level of certainty, while the latter exhibits relationships between variables. The two types of representations are semantically equivalent when they lead to the same possibility distribution (which rank-orders the possible interpretations). A possibility distribution can be decomposed using a chain rule which may be based on two different kinds of conditioning which exist in possibility theory (one based on product in a numerical setting, one based on minimum operation in a qualitative setting). These two types of conditioning induce two kinds of possibilistic graphs. In both cases, a translation of these graphs into possibilistic bases is provided. The converse translation from a possibilistic knowledge base into a min-based graph is also described.

## 1 Introduction

Possibilistic logic is issued from Zadeh's possibility theory [16], which offers a framework for the representation of states of partial ignorance owing to the use of a dual pair of possibility and necessity measures. Possibility theory may be quantitative or qualitative according to the range of these measures which may be the real interval $[0,1]$ or a finite linearly ordered scale as well [8]. Possibilistic logic (e.g., [7]) has been developed for more than ten years. It provides a sound and complete machinery for handling qualitative uncertainty with respect to a semantics expressed by means of possibility distributions which rank-orders the possible interpretations. At the syntactic level, possibilistic logic handles pairs of the form $(p\ \alpha)$ where $p$ is a classical propositional logic formula and $\alpha$ is an element of a totally ordered set. The pair $(p\ \alpha)$ expresses that the formula $p$ is certain at least to the level $\alpha$, or more formally by $N(p) \geq \alpha$, where $N$ is the necessity measure associated to the possibility distribution expressing the underlying semantics. Possibilistic logic is essentially *qualitative* since only the preordering induced on the formulas is important ($N(p) > N(q)$ means "p is more certain than q"). Possibilistic logic is as tractable as classical logic since its complexity is about $\log_2 n*SAT$ where $n$ is the number of certainty levels used in the knowledge base and SAT is the complexity of satisfaction problem in classical logic. Besides, there exist few works on directed possibilistic graphs (which are the counterpart of Bayesian probabilistic networks [13,14] in the framework of possibility theory). Existing works are either a direct adaptation of a probabilistic logic approach without caring for knowledge representation [10], or a way to do learning from imprecise data [11]. Because of the existence of the possibilistic logic machinery, there is not the same necessity to introduce graphical structures in possibility theory, as in probability theory (where probabilistic logic is more complex to handle). Yet Bayesian-like networks have a clear appeal for knowledge acquisition and directed graphs could be used to help in the specification of possibilistic knowledge as much as probabilistic knowledge.

Next section gives a background on possibilistic logic, on conditioning in possibility theory and on the directed possibilistic graphs. Section 3 studies their encoding in possibilistic logic. Section 4 discusses the problem of recovering the initial conditional possibility distribution from the joint possibility distribution computed with the chain rule, and the discussion briefly refers to the idea of possibilistic independence. Section 5 proposes an encoding of a set of possibilistic logic formulas into directed possibilistic graphs. Proofs are omitted for the sake of brevity, but can be found in [2].

## 2 Background

### 2.1. Possibilistic knowledge bases

A possibilistic (knowledge or belief) base is a set of possibilistic logic formulas of the form $(p,\alpha)$ where $p$ is a classical propositional logic formula; $\alpha$ an element of the semi-open real interval $(0,1]$ in a numerical setting, or of a finite linearly ordered scale in a qualitative setting. It estimates to what extent it is certain that $p$ is true considering the available, possibly incomplete information about the world.

Given a possibilistic base $\Sigma$, we can generate a possibility distribution from $\Sigma$ by associating to each classical interpretation a degree in $[0,1]$ expressing the level of compatibility with the available information. When a possibilistic base is made of one formula $\{(p\ \alpha)\}$, then each interpretation $\omega$ which satisfies p gets the degree $\pi(\omega) = 1$ since it is completely consistent with p, and each interpretation $\omega$ which falsifies p gets a degree $\pi(\omega)$ such that the highest is $\alpha$ (i.e., the more certain is p), the lowest is $\pi(\omega)$. In particular, if $\alpha$=1 (i.e., p is completely certain), then $\pi(\omega) = 0$, i.e., $\omega$ is impossible. One way to realize this constraint is to assign to $\pi(\omega)$ the degree 1 - $\alpha$ (on an ordered scale, we use a reversing map of the scale). Then, the possibility distribution associated to $\{(p\ \alpha)\}$ is:

$$\forall \omega \in \Omega,\ \pi_{\{(p\ \alpha)\}}(\omega) = 1 \quad \text{if } \omega \models p$$
$$= 1 - \alpha \quad \text{otherwise.}$$

When $\Sigma = \{(p_i, \alpha_i),\ i=1,n\}$ is a general possibilistic base then all the interpretations satisfying all the beliefs in $\Sigma$ will have the highest possibility degree, namely 1, and the other interpretations are ranked w.r.t. the highest belief that they falsified, namely we get $\forall \omega \in \Omega$:

$$\pi_\Sigma(\omega) = 1 \quad \text{if } \omega \models \Sigma$$
$$= 1 - \max\{\alpha_i : (p_i\ \alpha_i) \in \Sigma \text{ and } \omega \models \neg p_i\} \quad \text{otherwise.}$$

Thus, $\pi_\Sigma$ can be viewed as the result of the combination of the $\pi_{\{(p_i\ \alpha_i)\}}$'s using the min operator, i.e.:

$$\pi_\Sigma(\omega) = \min\{\pi_{\{(p_i\ \alpha_i)\}}(\omega) : (p_i\ \alpha_i) \in \Sigma\ \}.$$

A possibility distribution $\pi_\Sigma$ is said to be *normal* if there exists an interpretation $\omega$ which is totally possible, namely $\pi_\Sigma(\omega) = 1$. However, in general there may exist several distinct interpretations which are totally possible. This *normalization condition* reflects the consistency of the available knowledge $\Sigma$ represented by this possibility distribution (i.e., $\exists \omega \in \Omega$, s.t., $\omega \models \Sigma$). A possibility distribution $\pi$ induces two mappings grading respectively the possibility and the certainty of a formula p:

– the possibility degree $\Pi(p) = \max\{\pi(\omega) : \omega \models p\}$ which evaluates to what extent p is consistent with the available knowledge expressed by $\pi$. Note that we have:

$$\forall p\ \forall q \quad \Pi(p \vee q) = \max(\Pi(p), \Pi(q));$$

– the necessity (or certainty) degree $N(p) = min\{1 - \pi(\omega) : \omega \models \neg p\}$ which evaluates to what extent p is entailed by the available knowledge. We have:

$$\forall p\ \forall q \quad N(p \wedge q) = min(N(p), N(q)).$$

It can be checked that if $(p,\alpha) \in \Sigma$, the semantic constraint $N(p) \geq \alpha$ holds, where N is a necessity measure based on $\pi_\Sigma$. Note the duality equation: $N(p) = 1 - \Pi(\neg p)$. Moreover, note that, contrasting with probability theory $N(p)$ and $N(\neg p)$ (resp. $\Pi(p)$ and $\Pi(\neg p)$) are not functionally related: we only have (for normal possibility distributions) $min(N(p), N(\neg p)) = 0$ (resp. $max(\Pi(p), \Pi(\neg p)) = 1$). It leaves room for representing complete ignorance in an unbiased way: p is ignored whenever $\Pi(p) = \Pi(\neg p)) = 1$.

Lastly, several syntactically different possibilistic belief bases may have the same possibility distribution as a semantic counterpart. In such a case, it can be shown that these bases are equivalent in the following sense: their $\alpha$-cuts, which are classical bases, are logically equivalent in the usual sense, where the $\alpha$-cut of a possibilistic base $\Sigma$ is the set of classical formulas whose level of certainty is greater than or equal to $\alpha$.

## 2.2. Possibilistic graph

Let $V = \{A_1,...,A_n\}$ be a set of variables. These variables will identify the nodes of a network. In possibilistic logic, these variables are binary. The domains associated with the variables $A_i$ are denoted by $D_V = \{D_1,...,D_n\}$. A possible assignment of a value of $D_i$ to every variable $A_i$ will be called an elementary event. When each variable is binary (i.e., $D_i = \{a_i, \neg a_i\}$), the elementary events are also called interpretations and are denoted by $\omega$. The set of all the elementary events is simply the Cartesian product $D_1 \times ... \times D_n$ of the domains. Formulas, represented as sets of interpretations, are also called events.

Possibilistic graphs, denoted by $\Pi G$, as probabilistic networks, are based on directed acyclic graph [10, 11]. The nodes represent variables (for example, the temperature of a patient, the colour of a car, …) and the edges encode the causal link (or influence) between these variables. Uncertainty is represented on each node. When there is an edge from the node $A_i$ to the node $A_j$, the node $A_i$ is said to be "parent" of $A_j$ (parents of A are denoted by $Par(A)$). A possibility measure $\Pi$ is associated with a graph G in the following way:

• For the nodes $A_i$ which are roots of the graph (i.e., $Par(A_i) = \varnothing$), we specify the prior possibility degrees associated with each instance of $A_i$, namely we give every $\Pi(a)$ where $a \in D_i$. Possibilities must satisfy the normalization condition: $\max_{a \in D_i} \Pi(a) = 1$.

• To each other node $A_j$ are attached conditional possibilities $\Pi(a \mid \omega_{Par(A_j)})$ where $a \in D_j$ et $\omega_{Par(A_j)}$ is an element of the Cartesian product $\times_i D_i$ of the $D_i \in Par(A_j)$, domains associated with the variables of $Par(A_j)$. Conditional possibilities should satisfy the following normalization condition:

$$\forall \omega_{Par(A)} \in \times_i D_i,\ \max_{a \in D_j} \Pi(a \mid \omega_{Par(A)}) = 1,$$

$\Pi(.|b)$ is a conditional possibility measure, which can be defined in two different ways (as explained after the following Definition). Then:

**Definition 1**: The joint possibility distribution associated with a possibilistic graph $\Pi G$ is computed with the following equation:

$$\pi^\square_{\Pi G}(x_1...x_n) = \square_{i=1,n}\ \Pi(x_i \mid \omega_{x_i}) \qquad (1),$$

where $x_i \in \{a_i, \neg a_i\}$ are the two possible instances of the variable $A_i$, $\omega_{x_i} \subseteq \{x_1...x_n\}$ and $\square$ represents either the minimum or the product. We shall use $\pi_*$ and $\pi_m$ for short, in a case $\square$ = product or $\square$ = min respectively.

Indeed, in possibility theory, there exist two definitions for the conditioning:

• In a qualitative setting the conditioned possibility measure is defined by:

$$\Pi(q|p) = 1 \quad \text{if } \Pi(p \wedge q) = \Pi(p)$$
$$= \Pi(p \wedge q) \text{ otherwise (i.e., } \Pi(p \wedge q) < \Pi(p)),$$

and obeys the following equation [12] :

$$\Pi(p \wedge q) = min(\Pi(q|p), \Pi(p)); \qquad (2)$$

This definition of conditioning only requires the ordering between interpretations, and can be defined on any finite ordeed scale.

• In a numerical setting we use the following definition of conditioning based on the product :

$$\Pi(p \wedge q) \quad = \Pi(q|p) * \Pi(p) \quad \text{if } \Pi(p) \neq 0$$
$$= 0 \quad \text{otherwise.} \quad (3)$$

In both cases we have $N(q|p) = 1 - \Pi(\neg q|p)$. Up to a rescaling, (3) is also the conditioning rule of kappa-functions [15]:

$$\kappa(\omega|p) = \kappa(\omega) - \kappa(p) \quad \text{if } \omega \models p$$
$$= \infty \quad \text{otherwise,}$$

with $\Pi(\phi) = 2^{-\kappa(\phi)}$. Conditioning a possibility distribution with p then with r gives the same result as conditioning with r and then with p.

When the joint possibility distribution (1) is computed with the minimum (resp. product), the $\Pi G$ will be denoted by $\Pi G_{min}$ (resp. $\Pi G*$).

### 2.3. Decomposition

The decomposition of a possibility distribution consists in expressing a joint possibility distribution as a combination of conditional possibility distributions. For this decomposition, we can follow the same way as in probability theory. Let $\{A_1,...,A_n\}$ be the set of variables which is ordered arbitrarily. From the definition of conditioning, we have :

$$\pi(A_1...A_n) = \min [\pi(A_n|A_1...A_{n-1}), \pi(A_1...A_{n-1})].$$

Applying repeatedly this definition to $\pi(A_1...A_{n-1})$, then to $\pi(A_1...A_{n-2})$ ..., the joint possibility distribution is decomposed into

$$\pi(A_1...A_n) = \min [\pi(A_n|A_1...A_{n-1}),..., \pi(A_1)]. \quad (4)$$

The decomposition given by equation (4) can be simplified by assuming conditional independence between variables. When conditioning with the product, the decomposition follows the way used in probability:

$$\pi(A_1...A_n) = \pi(A_n|A_1...A_{n-1})*...*\pi(A_1). \quad (5)$$

There is no unique decomposition of a possibility distribution since it depends on the initial ordering between variables as it is shown in the following example:

**Example 1:** Let $\pi$ be defined on $\{a,\neg a\} \times \{b, \neg b\}$:
$\pi(\neg a \neg b)=1$; $\pi(a b) = \pi(a \neg b) =.8$; $\pi(\neg a b) = .7$.
There are two ways for decomposing this possibility distribution,

• either $\pi(A, B) = \min (\pi(B|A), \pi(A))$, and hence $\pi(B|A)$ is given by the matrix:

| B\A | a | ¬a |
|-----|---|-----|
| b | 1 | .7 |
| ¬b | 1 | 1 |

and $\pi(A)$ by the values $\pi(a)=.8$ and $\pi(\neg a)=1$,

• or $\pi(A, B) = \min (\pi(A|B), \pi(B))$, and hence $\pi(A|B)$ is given by the matrix:

| A\B | b | ¬b |
|-----|---|-----|
| a | 1 | .8 |
| ¬a | .7 | 1 |

and $\pi(B)$ by the values $\pi(b)=.8$ and $\pi(\neg b)=1$.

It can be verified that the two possible decompositions of the possibility distribution $\pi$ correspond to the two following possibilistic knowledge bases:

$\Sigma_1 = \{(\neg b \vee a \ .3), (\neg a \ .2)\},$
$\Sigma_2 = \{(b \vee \neg a \ .2), (\neg b \vee a \ .3), (\neg b \ .2)\}.$

These two bases (which can be computed as explained in next sections) are equivalent since they are associated with the same possibility distribution (i.e., $\pi_{\Sigma 1} = \pi_{\Sigma 2}$).

## 3. From the graph to the logical base

The goal of this section is to translate a directed possibilistic graph into a possibilistic base. The translation should be such that the possibility distribution associated to the graph using the chain rule is the same as the one associated to the possibilistic knowledge base. The restriction to binary variables is made for the sake of simplicity. See [1] for an extension to the non-binary case.

The directed possibilistic graph can be seen as the result of the fusion of one-formula knowledge bases, each one containing a single possibilistic formula. Each formula corresponds to a conditional possibility of the directed possibilistic graph. A possibilistic causal network is viewed as a set of triples:

$\Pi G = \{(a, P_a, \alpha) : \Pi(a|P_a)=\alpha \neq 1 \text{ is an element of the}$
                              directed graph$\}$

where 'a' is an instance of the variable A and $P_a$ is an element of the Cartesian product $\times_i D_i$ of the $D_i$. It can be restricted to the conditional possibilities different from 1 since the ones which are equal to 1 are not used in the computation of joint possibility distributions.

### 3.1  Encoding $\Pi G_{min}$

Let us start with min-based conditioning. With each triple $(a, P_a, \alpha)$ of the directed possibilistic graph is associated the single formula $(\neg a \vee \neg P_a \ 1-\alpha)$. The conditional symbol $\neg a | P_a$ is encoded as the material implication $\neg a \vee \neg P_a$, using the duality $\Pi(a|P_a)=1-N(\neg a|P_a)$.

The joint possibility distribution obtained from a directed possibilistic graph using (1) is equivalent to the one obtained by min-combination of the possibility distributions associated to the formulas encoding the different triples of the directed graph:

**Proposition 1:** Let $\pi_{a_i}$ be the possibility distribution associated with the formula corresponding to the triple $(a_i, P_{a_i}, \alpha_i)$. Then: $\pi_m = \min_i \pi_{a_i}$.

As already suggested in the background section, the union of two possibilistic bases $\Sigma_1$ and $\Sigma_2$ corresponds to the min-combination of the two possibility distributions $\pi_1$ and $\pi_2$ associated with $\Sigma_1$ and $\Sigma_2$ respectively. Therefore, the following Proposition states the knowledge base associated with a directed possibilistic graph:

**Proposition 2:** The possibilistic knowledge base associated with $\Pi G_{min}$ is:

$\Sigma = \{(\neg y_i \vee \neg P_{y_i} \ 1-\alpha_i) / (y_i, P_{y_i}, \alpha_i) \in \Pi G\}.$

This result is important since it implies that every result known for possibilistic logic can be applied to directed possibilistic graphs.

**Example 2** : Consider the following DAG :

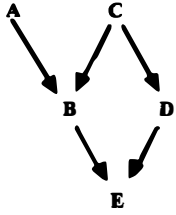

Assume that the conditional possibility degrees are given by the following tables :

Π(A)

| a | 1 |
|---|---|
| ¬a | 3/4 |

Π(C)

| c | 1 |
|---|---|
| ¬c | 1 |

Π(B | A C)

| B \| A C | a c | ¬a c | elsewhere |
|---|---|---|---|
| b | 1/2 | 1 | 1 |
| ¬b | 1 | 1/4 | 1 |

Π(D | C)

| D \| C | c | ¬c |
|---|---|---|
| d | 1 | 1/4 |
| ¬d | 1 | 1 |

Π(E | B D)

| E \| B D | b ¬d | ¬b d | elsewhere |
|---|---|---|---|
| e | 1 | 1/2 | 1 |
| ¬e | 3/4 | 1 | 1 |

We have:
RΠ = {(¬a, ∅, 3/4), (b, a c, 1/2), (¬b, ¬a c, 1/4), (d, ¬c, 1/4), (¬e, b ¬d, 3/4), (e, ¬b d, 1/2)}.
Then : $\Sigma$ = {(a 1/4), (¬b∨¬a∨¬c 3/4), (bva∨¬c 3/4), (¬d∨c 3/4), (e∨¬b∨d 1/4), (¬e∨b∨¬d 1/2)}.

### 3.2. Encoding ΠG*
Let us now turn towards product-based conditioning. Following the same approach as above, to each triple (a, $P_a$, α) of a possibilistic graph, is associated the single possibilistic formula (¬a∨¬$P_a$ 1-α). Let us notice that:
$$\Pi_a(a \mid P_a) = \Pi_a(a \wedge P_a) / \Pi_a(P_a) = \alpha,$$
where $\Pi_a$ is the possibility measure obtained from the possibility distribution associated with (¬a∨¬$P_a$ 1-α). Indeed, $\Pi_a(a \wedge P_a) = \alpha$ since each interpretation satisfying a∧$P_a$ falsifies (¬a∨¬$P_a$ 1-α). Moreover, $\Pi_a(P_a) = \max(\Pi_a(a \wedge P_a), \Pi_a(\neg a \wedge P_a))$. As every interpretation satisfying ¬a∧$P_a$ satisfies (¬a∨¬$P_a$ 1-α), we get $\Pi_a(P_a)=1$. The counterpart of Proposition 1 holds:

**Proposition 3:** Let $\pi_{ai}$ be the possibility distribution associated with (ai, $P_{ai}$, $\alpha_i$). Then $\pi* = *_i \pi_{ai}$, where * is the pointwise product of possibility distributions.

We need to find what is the possibilistic base associated with the combination of two possibility distributions by the product operator. This is summarized in the following proposition [3]:

**Proposition 4:** Let $\Sigma_1$={(pi αi) : i∈I} and $\Sigma_2$={(qj βj) : j∈J}. Let $\pi_1$ and $\pi_2$ be the two possibilistic distributions associated with $\Sigma_1$ and $\Sigma_2$. Let $\pi*$ be the product-based combination of $\pi_1$ and $\pi_2$. The resulting base associated with $\pi*$ is:
$$\mho^*(\Sigma_1, \Sigma_2) = \Sigma_1 \cup \Sigma_2 \cup$$
$$\{(pi∨qj (\alpha_i+\beta_j-\alpha_i*\beta_j)) : (pi \alpha_i)∈\Sigma_1 \text{ and } (qj \beta_j)∈\Sigma_2\}.$$

Since the operator $\mho*$ is commutative and associative, this straithforwardly extends to the case of more than two bases. Hence, the possibilistic base which is associated to ΠG* can be obtained by combining the different triples of the graph with the syntactic operator $\mho*$.

**Example 2** (continued) : Recall that :
RΠ={(¬a, ∅, 3/4), (b, a c, 1/2), (¬b, ¬a c, 1/4), (d, ¬c, 1/4), (¬e, b ¬d, 3/4), (e,¬b d,1/2)}.
The six elementary knowledge bases associated to the triples of ΠG* are:

$\Sigma_1$={(a 1/4)} ;     $\Sigma_2$={(¬a∨¬b∨¬c 1/2)} ;
$\Sigma_3$={(a∨b∨¬c 3/4)} ;     $\Sigma_4$={(c∨¬d 3/4)} ;
$\Sigma_5$={(¬b∨d∨e 1/4)} ;     $\Sigma_6$={(b∨¬d∨¬e 1/2)}.

Now let us combine them with $\mho*$ :
- the combination of $\Sigma_1$ et $\Sigma_2$ leads to:
$\Sigma_{12}$=$\mho*(\Sigma_1, \Sigma_2)$ = {(a 1/4), (¬a∨¬b∨¬c 1/2)} ;

- combining the result with $\Sigma_3$ :
$\Sigma_{123}$=$\mho*(\Sigma_{12}, \Sigma_3)$
= {(a 1/4), (¬a∨¬b∨¬c 1/2), (a∨b∨¬c 13/16)} ;

- combining the result with $\Sigma_4$ :
$\Sigma_{1234}$=$\mho*(\Sigma_{123}, \Sigma_4)$
= {(a 1/4), (¬a∨¬b∨¬c 1/2), (a∨b∨¬c 13/16), (c∨¬d 3/4), (a∨c∨¬d 13/16)} ;

- combining the result with $\Sigma_5$ :
$\Sigma_{12345}$=$\mho*(\Sigma_{1234}, \Sigma_5)$
= {(a 1/4), (¬a∨¬b∨¬c 1/2), (a∨b∨¬c 13/16), (c∨¬d 3/4), (a∨c∨¬d 13/16), (¬b∨d∨e 1/4), (a∨¬b∨d∨e 7/16), (¬a∨¬b∨¬c∨d∨e 5/8)} ;

- combining the result with $\Sigma_6$ leads to the final base :
$\Sigma*$=$\mho*(\Sigma_{12345}, \Sigma_6)$
= {(a 1/4), (¬a∨¬b∨¬c 1/2), (a∨b∨¬c 13/16), (c∨¬d 3/4), (a∨c∨¬d 13/16), (¬b∨d∨e 1/4), (a∨¬b∨d∨e 7/16), (¬a∨¬b∨¬c∨d∨e 5/8), (b∨¬d∨¬e 1/2)} ∪
{(a∨b∨¬d∨¬e 5/8), (a∨b∨¬c∨d∨¬e 29/32), (b∨c∨¬d∨¬e 7/8), (a∨b∨c∨¬d∨¬e 29/32)}.

This knowledge base contains 13 clauses while in the case of min we only have 6. This clearly illustrates that the combination with the product leads to a larger knowledge base than if we combined with the minimum, due to Proposition 4. This comes with the fact that the product is not compatible with a finite scale.

## 4. Recovering initial data

A natural question when we compute the joint possibility distribution using the chain rule is to see if we recover the a priori and conditional probabilities given by the expert. In the probability theory the answer is always yes. The following proposition shows that this is also the case if the chaining rule is based on the product:

**Proposition 5:** Let $\Pi(a|Pa)$ be the conditional possibility distributions over the variables A in the $\Pi G^*$. Let $\pi_*$ be the joint possibility distribution obtained using the chain rule with the product (1), and its associated possibility measure $\Pi_*$. Then, for any conditional possibility distribution we have:

$$\Pi_*(a|Pa) = \Pi(a|Pa).$$

Concerning the minimum-based conditioning the answer is no, as it is illustrated by the following small example:

**Example 3** Let us consider the following graph:

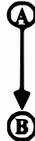

with $\Pi(a)=1$; $\Pi(\neg a)=1/4$; $\Pi(b|a) = 1/3$; $\Pi(\neg b|a) = 1$ and $\Pi(b|\neg a) = 1/3$; $\Pi(\neg b|\neg a) = 1$.
The joint possibility distribution is:

$\pi_m(ab) = 1/3$;         $\pi_m(a\neg b) = 1$;
$\pi_m(\neg ab) = 1/4$;      $\pi_m(\neg a\neg b) = 1/4$;

Clearly, we have $\Pi_m(b|\neg a) = 1 \neq 1$ since $\pi_m(\neg ab) = \pi_m(\neg a) = 1/4$.

The reason for not recovering the original values is that the conditional possibility distributions specified by the users are not coherent with the axioms of possibility distributions. Indeed, using the definition of conditional possibility measure, we always have:

If $\Pi(p|q) \neq 1$ then $\Pi(p|q) = \Pi(p \wedge q) < \Pi(q)$.

We see clearly, from the previous example that this constraint is violated since $\Pi(b|\neg a) = 1/3 \neq 1$ and $\Pi(b|\neg a) > \Pi(\neg a) = 1/4$. Therefore, it is not surprizing if we do not recover the above value.

This behaviour also exists in possibilistic logic, namely a possibility distribution associated a possibilistic base do not guarantee to recover the exact value of the knowledge base. To be convinced, just consider a small example where $\Sigma = \{(a\ .8), (a \vee b\ .4)\}$. We can easily check that $N_{\pi\Sigma}(a \vee b) = .8$. This is due to the fact that $(a \vee b\ .4)$ is strictly subsumed by $(a\ .4)$. Hence, we have:

**Proposition 6:** Let $\Sigma$ be such that it does not contain any strictly subsumed formulas, namely there is no $(p, \alpha)$ such that $\{q : (q, \beta) \in \Sigma \text{ and } \beta > \alpha\} \vdash p$. Then $\forall(p, \alpha) \in \Sigma$ we have: $N_{\pi\Sigma}(p) = \alpha$.

Let us go back to the causal network, the following proposition characterizes to what the computed joint possibility corresponds w.r.t. the conditional possibility distributions specified by the user.

**Proposition 7:** Let $\Pi(a|Pa)$ be the conditional possibility distributions over the variables A in the DAG. Let $\pi_m$ be the joint possibility distribution obtained using the chain rule with the minimum-based conditioning. Then: either $\pi_m(a|Pa) = \Pi(a|Pa)$ or $\pi_m(a|Pa) = 1$.

This means that the computed joint possibility distribution either preserves the initial values or push them up to 1 (this is observed in Example 1).
We now focus on the DAG, from where we recover the original values. In the possibility theory, the definition of independence is not unique [4][6]. The following is a weaker one called "non-interactivity":

**Definition 2:** Two variables A and B are independent in the context C, if for each instance (a,b,c) of (A, B, C) we have:    $\Pi(a,b\ |c) = \min(\Pi(a|c), \Pi(b|c))$.

The following proposition shows that the joint possibility distribution guarantees the independence relations from the structure of the DAG, as in probabilistic network:

**Proposition 8:** Let X be a given variable, and Y be a variable which is neither a parent of X nor any of its descendant. Let $\pi_m$ be the joint possibility distribution computed from a DAG $\Pi G$ using the min-based chain rule. Then X and Y are independent in the context of Par(X) in the sense of Definition 2.

The question of whether a joint possibility distribution can be decomposed using a stronger definition of independence is left for further research.

## 5. Encoding bases into min-based graphs

In this section, we present the transformation of possibilistic knowledge bases into directed possibilistic graphs $\Pi G_{min}$. One way to do it is to use possibility distributions as intermediary. Indeed, a knowledge base leads to a possibility distribution, from which it is possible to build a $\Pi G_{min}$ (see Section 2.3). This would apply as well to $\Pi G_*$. However, this way is computationally expensive. Moreover, we want to find the $\Pi G_{min}$ directly from the knowledge base.
The encoding of a possibilistic knowledge into a $\Pi G_{min}$ is less straightforward than the previous transformations. Indeed, we cannot directly view each formula as a triple and then build the graph, but we need some pre-processing steps. The constructed possibilistic graph $\Pi G$ should be such that:

• the joint possibility distribution computed from the $\Pi G$ using the minimum operator should be the same as the one computed from the knowledge base;

• the joint possibility distribution allows us to recover all the conditional possibilities of the DAG, and
• the joint possibility distribution satisfies the independence relation (in the sense of Definition 2) induced by the structure of the DAG, namely each variable should be independent of any, non-descendent, variable in the context of its parents.

The construction of a causal network associated to a possibilistic knowledge is obtained in three steps: the first step simply consists in putting the knowledge base into a clausal form and in removing tautologies. The second step consists in constructing the graph, and the last step computes the conditional possibilities associated to the constructed graph.

## 5.1 Putting bases in a clausal form
In this first step, a base $\Sigma$ is rewritten into an equivalent base $\Sigma'$. The equivalence means here: $\pi_\Sigma = \pi_{\Sigma'}$. Getting $\Sigma'$ consists in putting the knowledge base into a clausal form and in removing tautologies. The following proposition shows how to put the base in a clausal form first:

**Proposition 9** : Let $(p \; \alpha) \in \Sigma$. Let $\{c_1, ..., c_n\}$ be the set of clauses encoding p in classical logic. Let $\Sigma^*$ be a new knowledge base obtained from $\Sigma$ by replacing $(p \; \alpha)$ by $\{(c_1 \; \alpha), ..., (c_n \; \alpha)\}$. Then the two knowledge bases $\Sigma$ and $\Sigma^*$ are equivalent in the sense that $\pi_\Sigma = \pi_{\Sigma^*}$.

Then removing tautologies leads still to an equivalent possibilistic base. Indeed, tautologies are always satisfied, and $\pi_\Sigma(\omega)$ is only defined with respect to that formulas of $\Sigma$ falsified by $\omega$. The removing of tautologies is an important point since this will avoid having links which do not make sense. For example, the tautological formula $(\neg x \lor \neg y \lor x \; 1)$ could induce a link between X and Y.

## 5.2 Constructing the graph
The second step consists in constructing the graph, namely the determination of the vertices (the set of variables) of the graph and the parents of each vertex. The set of variables is simply the set of propositional symbols which appear in the knowledge base. Moreover, since possibilistic logic is based on propositional logic, then all the used variables are binary variables. By X we denote the variable which can be either x or $\neg x$. To construct the graph, we first rank the variables, according to an arbitrary numbering $\{X_1, X_2, ..., X_n\}$ of the variables. This ranking intends to mean that parents of each variable $X_1$ can only be in $\{X_{i+1}, ..., X_n\}$ (but they may not exist). We first give several intuitive examples before presenting the technical construction of the graph.

**Example 4**: Let: $\Sigma = \{(t \; 0.6), (t \lor v \; 0.4)\}$. From this knowledge base one may think that the variable T depends on the variable V. However, we can easily check that $\Sigma$ is equivalent to the following one: $\Sigma' = \{(t \; 0.6))\}$, where clearly, V has no influence on T. The formula $(t \lor v \; 0.4)$ is said to be subsumed by $\Sigma - \{(t \lor v \; 0.4)\}$.

Next definition formally introduces the notion of

subsumed beliefs, which can be removed as stated by Proposition 10:

**Definition 3**: A formula $(p \; \alpha)$ of $\Sigma$ is said to be subsumed if $\Sigma_{\geq \alpha} \models p$ where $\Sigma_{\geq \alpha} = \{q : (q \; \beta) \in \Sigma$, and $\beta \geq \alpha\} - \{(p \; \alpha)\}$ (namely p can be recovered from clauses of the base having weights greater or equal than $\alpha$).

**Proposition 10**: Let $\Sigma'$ be a new base obtained from $\Sigma$ by removing subsumed formulas. Then $\pi_\Sigma = \pi_{\Sigma'}$.

Therefore, we can remove, or add, subsumed beliefs without any damage. Subsumed beliefs are not the only ones which may induce fictitious dependencies:

**Example 5**: Let: $\Sigma = \{(a \; .5), (\neg a \lor b \; .5)\}$.
In this base, neither $(a \; .5)$ nor $(\neg a \lor b \; .5)$ is subsumed, and one can think that there is a relationship between the variables A and B. However, we can easily check that this base is equivalent to $\Sigma' = \{(a \; .5), (b \; .5)\}$, where clearly A and B are unrelated. So we are let to state:

**Proposition 11**: Let X be a variable, and $(x \lor p \; \alpha)$ be a clause of $\Sigma$ containing the instance x of X s.t. $\Sigma \vdash (p \; \alpha)$[1]. Then the base $\Sigma$ and the knowledge base $\Sigma'$ obtained from $\Sigma$ by replacing $(x \lor p \; \alpha)$ by $(p \; \alpha)$ are equivalent.

Intuitively, one can say that two variables are related if there is a clause containing an instance of these two variables, and they are unrelated otherwise. Example 6 shows that two variables can be related even if there is no clause in the base containing an instance of variables:

**Example 6**: Let: $\Sigma = \{(x \lor a \; .5), (\neg x \lor b \; .5)\}$.
In this base, if we would define the relationship between variables only if there exists a clause containing an instance of each of them, then clearly A and B would be unrelated. However, we can check that: $\Sigma \vdash (a \lor b \; .5)$.

The following example shows that in order to compute the conditional possibility distribution of a variable given its parents, it is not enough to look only to clauses of the base containing instances of the variable X:

**Example 7**: Let: $\Sigma = \{(a \lor b \; .6), (x \lor b \; .5), (x \lor a \; .5)\}$.
We assume that parents of X are A and B. Clearly, if we compute the conditional possibilities $\Pi(X|AB)$ only from clauses containing X, namely $\Sigma_X = \{(x \lor b \; .5), (x \lor a \; .5)\}$ then it is not guaranteed to recover all original values. Indeed, in this example, if the computation of $\Pi(\neg x|\neg a \neg b)$ is simply based on $\Sigma_X$ we get: $\Pi(\neg x|\neg a \neg b) = .5$

   (since $\{(x \lor b \; .5), (x \lor a \; .5)\} \vdash (x \lor a \lor b \; .5)$)
but we can check that after computing the joint possibility distribution $\pi_\Sigma$: $\Pi_\Sigma(\neg x|\neg a \neg b) = 1$, this is due to the fact that we have both: $\Sigma \vdash (x \lor a \lor b \; .6)$ and $\Sigma \vdash (a \lor b \; .6)$, hence: $\Pi_\Sigma(\neg x \land \neg a \neg b) = \Pi_\Sigma(\neg a \neg b)$.
Indeed, $(x \lor b \; .5)$ and $(x \lor a \; .5)$ are not subsumed but $(x \lor b \lor b$

---

.5) is because of the clause (a∨b .6) in Σ.

The following definition enables a direct computation of conditional possibility distributions:

**Definition 4:** Let X be a variable and Par (X) be the set of its parents. Let K be a set of clauses of the form x∨p such that P⊆Par (X) (P is the set of variables containing an instance in x∨p). We call complete extension of K, denoted by E(K), the set of all clauses of the form (x∨∨$_{y∈P_X}$ y, α) where x is an instance of X, $P_X$ is an instance of Par (X), and

α = max {a$_i$ : (x∨p$_i$  a$_i$) ∈ K and p$_i$⊨∨$_{y∈P_X}$ y},

with max {∅}=0.

**Example 8:** Let K = {(x∨b .5), (x∨a .5)}. Assume that the parents of X are A and B. Then we can check that: E(K) = {(x∨b∨¬a .5), (x∨b∨a .5), (x∨a∨¬b .5)}.

**Proposition 12:** The two bases K and E(K) are equivalent.

With the help of the previous propositions we are now ready to determine the exact parents of each variable. Let {X$_1$, …, X$_n$} be a numbering of the variables. Then the following algorithm determines the set of parents of each variable:

For i = 1,…, n do
Begin
        /* Determination of Parents of X$_i$ */
1. Let (x$_i$∨p, α) be a clause of Σ s. t. :
        x$_i$ is an instance of X$_i$, and
        p is only built from {X$_{i+1}$,…,X$_n$}.
  • If (x$_i$∨p, α) is subsumed, then remove it from Σ
        (due to Prop. 10).
  • If Σ ⊢ (p, α) then replace (x$_i$∨p, α) by (p, α)
        (due to Prop. 11)
2. Let K$_i$ be the set of clauses (x$_i$∨p, α) in Σ s. t.
        p is only built from {X$_{i+1}$,…,X$_n$}
3. The parents of the variable X$_i$ are :
        Par(X$_i$)= {X$_j$ : ∃ c ∈ K$_i$ such that c contains
        an instance of X$_j$}
        /* Rewriting Σ to facilitate the computation of
        conditional possibilities , and for looking for hidden
        dependencies */
4. Replace in Σ, K$_i$ by its complete extensions
        E(K$_i$) (due to Prop. 12).
5. For each (x$_i$∨p α) of Σ (where p is built from
        {X$_{i+1}$,…,X$_n$}) such that Σ ⊢ (p, α)
                replace (x$_i$∨p, α) by (p, α).
6. Let Σ$_{X_i}$ be the set of clauses (x$_i$∨p, α) in Σ s. t.
        p is only built from {X$_{i+1}$,…,X$_n$}
End.

The algorithm is composed of two parts: the first part (Steps 1-3) consists in the construction of the DAG by determining the parents of each variable, and the second part (Steps 4-6) consists in rewriting the knowledge base such that : i) it gives immediately the conditional possibility distributions, and ii) ensures to recover original values when using the chain rule for computing the joint possibility distributions. Indeed, once E(K$_i$) is computed, in order to evaluate Π(x|p$_X$) then either (¬x$_i$∨¬p, α)∉E(K$_{X_i}$) then Π(x|p$_X$) = 1, or (¬x$_i$∨¬p, α) ∈ E(K$_i$), then if Σ ⊢ (¬p, α) then Π(x|p$_X$) = 1 (since Π(x∧p$_X$) = Π(x)), otherwise Π(x|p$_X$) = 1 - α.

The result of the algorithm is a partition {Σ$_{X_1}$, …, Σ$_{X_n}$} such that Σ$_{X_1}$∪ …∪ Σ$_{X_n}$ is equivalent to Σ. Clearly, for i>1, Σ$_{X_i}$ does not contain any variable of {X$_1$, …, X$_{i-1}$}. Moreover, Σ$_{X_i}$ can be empty. In this case X$_i$ has no parents : it corresponds to the root of the graph, and the priori possibility degrees associated to X$_i$ are equal to 1.
The subbases Σ$_{X_i}$'s give a direct computation of conditional possibility degrees as explained later.

A graph associated to Σ is such that its vertices are the variables in Σ, and a link is drawn from X$_j$ to X$_i$ iff X$_j$ ∈ Par(X$_i$), where Par(X$_i$) is given by step 3 in the algorithm. This graph is indeed a DAG.

**Example 9 :** Let us consider the following base: Σ = {(a∨b 0.7), (¬a∨c∨¬d 0.7), (a∨c∨d 0.9), (b∨c 0.8), (¬b∨e 0.2), (¬d∨f 0.5)}
Σ contains six variables numbered in the following way: X$_1$=A, X$_2$=B, X$_3$=C, X$_4$=D, X$_5$=E et X$_6$=F.
From this partition we can check that we get:
Par(A)={B, C, D}, Par(B)={C, E}, Par(C)=∅, Par(D)={F}, Par(E)= PAR(F)=∅. The final graph is:

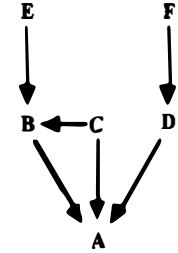

## 5.3 Determining the conditional possibilities

Once the graph is constructed we need to compute the conditional possibilities. The computation of Π(X$_i$| Par(X$_i$)) is immediately obtained from Σ$_{X_i}$, as it is given by the following way: Let {Σ$_{X_1}$, …, Σ$_{X_n}$} be the result of Step 6 of the previous algorithm. Let X$_i$ be a variable  Let x be an instance of X$_i$ and P$_X$=y$_1$∧…∧y$_m$ be an instance of Par (X$_i$). Let:
        Π(x|P$_X$) = 1−a$_i$ if (¬x ∨ ¬P$_X$, a$_i$) ∈ Σ$_{X_i}$
                    = 1 otherwise.
Then we have:

**Proposition 13:** The possibility distribution associated to Σ and the possibility distribution obtained from the graph using the minimum operator are equal, namely: π$_Σ$=π$_m$.

**Example 10** (continued) :
Let us show how compute $\pi(\neg a|\neg b\neg cd)$ and $\pi(\neg a|\neg b\neg c\neg d)$, the others are obtained in the same way. Recall that

$\Sigma_A$ = {(a∨b∨c∨d 0.7), (a∨b∨c∨¬d 0.7), (a∨b∨¬c∨d 0.7), (a∨b∨¬c∨¬d 0.7), (¬a∨b∨c∨¬d 0.7), (¬a∨¬b∨c∨¬d 0.7), (a∨b∨c∨d 0.9), (a∨¬b∨c∨d 0.9)}.

Then by definition:
$$\pi(\neg a|\neg b\neg cd) = .3$$
$$\pi(\neg a|\neg b\neg c\neg d) = .1$$

The conditional possibility distributions are given by the following tables:

$\pi(A \mid BCD)$ :

| A \| BCD | ¬b¬cd | ¬bc¬d | ¬bcd | ¬b¬c¬d |
|---|---|---|---|---|
| a | 1 | 1 | 1 | 0.3 |
| ¬a | 0.3 | 0.3 | 0.3 | 1 |

| A \| BCD | b¬cd | ¬b¬c¬d | b¬c¬d | otherwise |
|---|---|---|---|---|
| a | 0.3 | 1 | 1 | 1 |
| ¬a | 1 | 0.1 | 0.1 | 1 |

$\pi(B \mid CE)$ :

| B \| CE | ¬c¬e | ¬ce | ¬c¬e | c¬e |
|---|---|---|---|---|
| b | 1 | 1 | 0.8 | 0.8 |
| ¬b | 0.2 | 0.2 | 1 | 1 |

$\pi(D \mid \neg F)$ :

| D \| F | f | ¬f |
|---|---|---|
| d | 1 | 0.5 |
| ¬d | 1 | 1 |

$\pi(C)$ :

| | |
|---|---|
| c | 1 |
| ¬c | 1 |

$\pi(E)$ :

| | |
|---|---|
| e | 1 |
| ¬e | 1 |

$\pi(F)$ :

| | |
|---|---|
| f | 1 |
| ¬f | 1 |

Let us notice that it is possible to express ignorance, i.e., it is not necessary to give the a priori possibilities on X if they are unknown.

**Proposition 14:** Let $\Pi(X|Par(X))$ be the conditional possibility distributions over the variables X in the DAG associated to $\Sigma$. Let $\pi_m$ be the joint possibility distribution obtained using the chain rule with minimum from a causal network. Then: $\pi_m(a|Pa) = \Pi(a|P_a)$.

The previous proposition shows that the constructed DAG from the possibilistic logic base guarantees the recovering of initial values, when using the min-based chain rule, and hence it satisfies the independence relations in the sense of Definition 2 encoded by the structure of the graph.

## 6.   Conclusion

This paper has established the links between possibilistic logic and directed possibilistic graphs. We have shown that directed possibilistic graphs can be encoded into possibilistic logic, for the two possible definitions of conditioning based on the minimum and the product operator. When it is based on product, it also provides a mean for turning a Bayesian-net equipped with a kappa-function into a possibilistic logic base (using the transformation possibility measures-kappa functions recalled at the end of Section 2.2.).

The inverse passage from a possibilistic logic base to a network has also been provided. This allows the expert to express his knowledge using "causality" relations between variables, and then the possibilistic logic machinery can be applied after the computation of the corresponding possibilistic logic base.

A future work would be the study of the complexity of the conversion of a directed causal network based on the product into possibilistic logic (with the min, the conversion is linear) and the comparison of the cost of the inference using the network directly with the one using the corresponding possibilistic knowledge base, which may be also used for explanation purposes.